\pdfoutput=1
\documentclass[11pt]{article}

\usepackage[final]{acl}

\usepackage[utf8]{inputenc}

\usepackage{times}
\usepackage{latexsym}

\usepackage[T1]{fontenc}

\usepackage{microtype}

\usepackage{inconsolata}

\usepackage{graphicx}
\usepackage{tabularx}
\usepackage{multirow}
\usepackage{booktabs}
\usepackage{tcolorbox}
\usepackage{listings}
\usepackage{soul}

\usepackage{tikz}
\usetikzlibrary{shapes.geometric, arrows.meta, positioning, fit, backgrounds}

\lstnewenvironment{prompt}[1]
{%
    \lstset{
      caption=#1,
      captionpos=b,
      basicstyle=\ttfamily\small,
      columns=fullflexible,
      frame=single,
      breaklines=true,
      breakindent=0pt,
      backgroundcolor=\color{gray!10}
    }%
}{}

\usepackage{xcolor}

\usepackage{xspace}
\newcommand{\DCLM}{\textsc{DCLM}\xspace}
\newcommand{\Ours}{Nemotron-CC\xspace}

\title{Nemotron-CC: Transforming Common Crawl into a Refined Long-Horizon Pretraining Dataset}

\author{
 \textbf{Dan Su\textsuperscript{*}},
 \textbf{Kezhi Kong\textsuperscript{*}},
 \textbf{Ying Lin\textsuperscript{*}},
 \textbf{Joseph Jennings},
 \textbf{Brandon Norick},
\\
 \textbf{Markus Kliegl\textsuperscript{\textdagger}},
 \textbf{Mostofa Patwary},
 \textbf{Mohammad Shoeybi},
 \textbf{Bryan Catanzaro}
\\
\\
  NVIDIA
\\
 \small{
   \textsuperscript{*}Equal contribution. \textsuperscript{\textdagger}Correspondence to \href{mailto:mkliegl@nvidia.com}{mkliegl@nvidia.com}.
 }
}

\begin{document}
\maketitle
\begin{abstract}
Recent English Common Crawl datasets like FineWeb-Edu and \DCLM achieved significant benchmark gains via aggressive model-based filtering, but at the cost of removing $90\%$ of data. This limits their suitability for long token horizon training, such as 15T tokens for Llama 3.1. In this paper, we show how to achieve better trade-offs between accuracy and data quantity by a combination of classifier ensembling, synthetic data rephrasing, and reduced reliance on heuristic filters.
When training 8B parameter models for 1T tokens, using a high-quality subset of our data improves MMLU by 5.6 over \DCLM, demonstrating the efficacy of our methods for boosting accuracies over a relatively short token horizon. Furthermore, our full 6.3T token dataset matches \DCLM on MMLU, but contains four times more unique real tokens than \DCLM. This unlocks state-of-the-art training over a long token horizon: an 8B parameter model trained for 15T tokens, of which 7.2T came from our dataset, is better than the Llama 3.1 8B model: +5 on MMLU, +3.1 on ARC-Challenge, and +0.5 on average across ten diverse tasks.
The dataset is available at \url{https://data.commoncrawl.org/contrib/Nemotron/Nemotron-CC/index.html}.
\end{abstract}

\section{Introduction}

\begin{figure}[th!]
\centering
\includegraphics[width=0.8\linewidth]{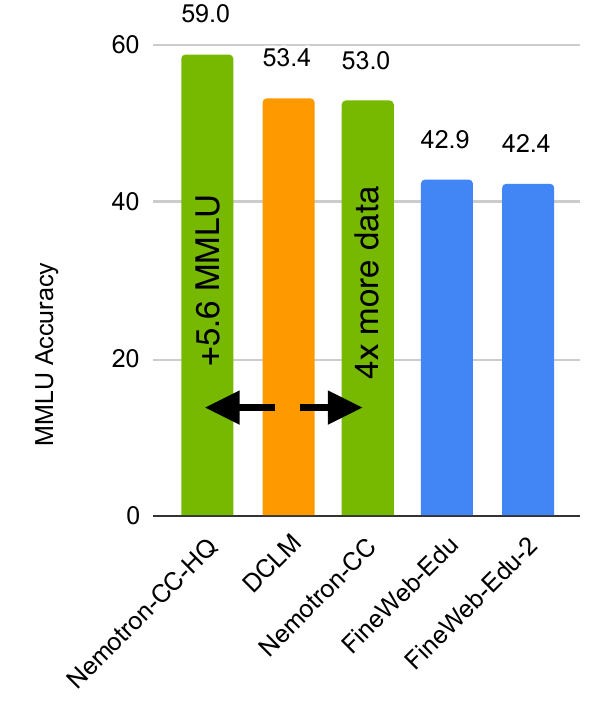}
\caption{MMLU scores for 8B parameter models trained for 1T tokens. Compared to \DCLM, our methods enable us to either create a 4$\times$ larger dataset of similar quality or increase the MMLU using a high quality subset of the tokens. Having a larger dataset, in the sense of unique real tokens, is crucial when training over long horizons such as 15T tokens.}
\label{fig:main}
\end{figure}

Internet crawl is the largest source of unique tokens for training LLMs and can be seen as serving two main purposes: high-quality content and diversity.  Recent English datasets derived from Common Crawl\footnote{\url{https://commoncrawl.org/}} such as FineWeb-Edu~\citep{penedo2024fineweb} and \DCLM~\citep{li2024datacomp} have emphasized high-quality content that boosts benchmark accuracies over data quantity. They have demonstrated significant strides in achieving benchmark results competitive with some of the best closed models at a small scale (e.g., DCLM's 7B model trained over 2.6T tokens), primarily thanks to the use of model-based filters to extract high-quality educational and instructional content. However, this comes at the cost of data quantity: they remove around $90\%$ of the data.
Such aggressive pruning may not be the most effective strategy when training larger models over longer token horizons (e.g., Llama 3.1 includes 8B--405B parameter models, trained for 15T tokens~\citep{dubey2024llama} and Gemma 2 27B was trained for 13T tokens~\citep{team2024gemma}).
Both \DCLM and FineWeb-Edu contain around $80\%$ near-duplicates (1T and 0.2T unique tokens, respectively)~\citep{fineweb-edu-discussion,li2024datacomp} and to train on these datasets for many trillions of tokens implies seeing essentially the same samples many times during training. This could lead to inferior models, as \citet{muennighoff2024scaling} find there are diminishing returns after four epochs compared to training on more unique tokens.

In this paper, we show how to achieve a better trade-off between benchmark accuracy and data quantity with a combination of classifier ensembling, synthetic data generation, and reduced reliance on heuristic filters. Our main contributions are:
\begin{enumerate}
\item We propose a method for transforming English Common Crawl into a 6.3T token long-horizon pretraining dataset, consisting of 4.4T globally deduplicated original tokens and 1.9T synthetically generated tokens.
We release the dataset\footnote{\url{https://data.commoncrawl.org/contrib/Nemotron/Nemotron-CC/index.html}} under the Common Crawl Terms of Use and a reference implementation as part of the Apache 2.0 open-source NeMo Curator library.\footnote{\url{https://github.com/NVIDIA/NeMo-Curator}} The quality classifier models have been released as well.\footnote{\url{https://huggingface.co/nvidia/nemocurator-fineweb-nemotron-4-edu-classifier} and \url{https://huggingface.co/nvidia/nemocurator-fineweb-mixtral-edu-classifier}}
\item We prove the effectiveness of this method by comparing to the state-of-the-art open English Common Crawl datasets \DCLM and FineWeb-Edu~(Figure~\ref{fig:main}).
\begin{enumerate}
\item A 1.1T-token high-quality subset of our data achieves a 5.6 MMLU improvement over \DCLM, showing the superiority of our method over a relatively short token horizon.
\item Our full dataset performs on par with \DCLM while having 4$\times$ as many unique real tokens.
\item This larger size enables state-of-the-art results over long token horizons: An 8B parameter model trained for 15T tokens using a weighted version of our dataset achieves higher overall accuracy than Llama 3.1 8B, and in particular MMLU 70.3 vs. Llama's 65.3. Note that Llama 3.1 8B was also trained on 15T tokens~\citep{dubey2024llama}.
\end{enumerate}
\item We conduct ablation studies and find:
\begin{enumerate}
\item Ensembling different model-based classifiers can help select a larger and more diverse set of high quality tokens.
\item Rephrasing can effectively reduce noise and errors in low-quality data and produce diverse variants with fresh unique tokens from high-quality data, leading to better results in downstream tasks.
\item Disabling traditional non-learned heuristic filters for high-quality data can further boost high quality token yield without hurting accuracy.
\end{enumerate}
\end{enumerate}

Finally, we remark that our overall guiding principle is to shift from a static, non-learned, heuristic pipeline towards a more learned flywheel whose performance will naturally get better over time. As our data improves, so will the LLMs we train, and these improved LLMs will in turn improve our data as we use them to generate better synthetic data and quality classifications.

\section{Methods}

In this section we explain our efforts to build the best English Common Crawl pretraining dataset for LLMs. Our efforts can be split into three folds. First, we talk about our efforts in boosting token yield by utilizing text extractor and heuristic filters more properly in Section~\ref{sec:extract_filter}. Second, we introduce the model-based quality labeling pipeline methods in Section~\ref{sec:classifier}. Third, we introduce our synthetic data generation method to further improve the data quality in Section~\ref{sec:synthetic}. For a schematic overview of our final pipeline, please see Figure~\ref{fig:pipelineoverview} in Appendix~\ref{appendix:pipelineoverview}.

\subsection{HTML-to-text Extractor \& Filter} \label{sec:extract_filter}
Extracted texts from HTMLs are the foundation and major source of LLM pretraining dataset, so it is of great significance to analyze and understand the extraction tools for optimal data quality and token yield. Moreover, heuristic filters are often utilized to remove low-quality tokens with human-designed heuristics \citep{li2024datacomp,parmar2024data,penedo2024fineweb,dubey2024llama}, which may also put good tokens at the risk of being removed. We carefully examine both aspects with the assist of the FineWeb-Edu classifier \citep{penedo2024fineweb}, a model-based quality classifier that had shown effectiveness in identifying high-quality tokens that are significant in boosting the strength of LLMs.

\begin{table}[htbp] \small \centering
\begin{tabular}{@{}lccc@{}}
\toprule  
                     & \textbf{\#Tokens} & \textbf{\#HQ tokens} & \textbf{\#HQ +\%} \\ \midrule
Trafilatura-filtered & 994              & 80                  & -                       \\
Justext-filtered     & 1,380            & 104                 & 28.6\%                  \\
Justext              & 1,804            & 127                 & 57.4\%                  \\ \bottomrule
\end{tabular}
\caption{Extraction and filteration token count statistics (billion). Tokens counted after deduplication.}
\label{table:extract_filter_stats}
\end{table}


\paragraph{HTML-to-text Extraction} We test two HTML-to-text extractors, Justext~\citep{pomikalek2011removing} and Trafilatura~\citep{barbaresi-2021-trafilatura}. Qualitatively, we view both extractors at the same level of quality. Quantitatively, we calculate token yields of both extractors on 13 selected snapshots of Common Crawl (see Appendix~\ref{sec:13_snapshots}). The statistics are reported in Table~\ref{table:extract_filter_stats}. We see that Justext can yield more tokens, notably more high-quality tokens (+28.6\%) by the standard of Fineweb-Edu classifier (score 3, 4, and 5). We highlight that boosting unique token amount is of great importance when building long-horizon pretraining dataset, e.g., 15T tokens for Llama3.1. Even though there is a slight decline in the percentage of HQ tokens for Justext vs. Justext-filtered ($7.0\%$ vs. $7.5\%$), what we aim to maximize here is the absolute number of HQ tokens (127B vs. 104B). We will later sort the data into quality buckets, which enables exact control of the proportion of HQ vs. non-HQ data seen during training instead of reliance on the natural distribution for a particular extraction tool.
After extraction, we apply filtering to keep only English text, as determined by pycld2\footnote{\url{https://pypi.org/project/pycld2/}} and the FastText lid176 language classifier\footnote{\url{https://fasttext.cc/docs/en/language-identification.html}} with threshold 0.3~\citep{joulin2016fasttext,joulin2016bag}.
We then apply global fuzzy deduplication as well as exact substring deduplication over eighths of snapshots~\citep{lee2022deduplicating}, using the NeMo Curator library\footnote{\url{https://github.com/NVIDIA/NeMo-Curator}} and the deduplicate-text-datasets library,\footnote{\url{https://github.com/google-research/deduplicate-text-datasets}} respectively.

\paragraph{Filtering} Conventionally, heuristic filters are leveraged to remove low-quality tokens from the pretraining dataset as a post-processing step \citep{li2024datacomp,parmar2024data,penedo2024fineweb,dubey2024llama}. We revisit the filtering pipeline as in \citep{parmar2024data}. Such pipeline sequentially consists of a set of heuristic filters proposed by \citet{raffel2020exploring,rae2021scaling} and a perplexity filter based on a KenLM model~\citep{heafield2011kenlm} trained on Wikipedia and books data~\citep{wenzek-etal-2020-ccnet}. To quantitatively better understand the effectiveness of the filtering pipeline, we calculate the token yield and report the numbers in Table \ref{table:extract_filter_stats}. We find the filtering pipeline removes a non-trivial portion of high-quality tokens (-18.1\%) classified by FineWeb-Edu classifier from the dataset. 

Given the impact that the heuristic filters have on the high-quality token yield, we propose to NOT apply such filters to the high-quality tokens distinguished by model-based quality classifers (described in the next section), but only use those on the low-quality splits. In the experiment section we empirically verify the impact of both the extractor and filter on pretraining data quality through downstream benchmarks. We refer readers to Section~\ref{sec:ablation} for detailed results.

\subsection{Model-based Quality Labeling} \label{sec:classifier}
Recent work~\citep{li2024datacomp,penedo2024fineweb} use model-based classifiers to extract high-quality pretraining documents from English Common Crawl. However, both of the two quality classifiers have a limited recall (around 10\%) of high-quality tokens (see Table~\ref{table:classifier_ablation}), and this will become a bottleneck to train an LLM over a long horizon. Also, the quality labels assigned by the quality classifier are not necessarily aligned with LLM's downstream task performance. Therefore, we propose our ensemble-based quality labeling pipeline method. Specifically, we first build three quality classifiers, each of which has different high-quality preferences. Then, we ensemble the three classifiers to score all the documents, and split the crawl corpus into different quality buckets based on the quality score. Finally, we regroup the fine-grained document buckets into 5 different quality levels based on their corresponding performance on downstream task. 

\paragraph{Quality Classifier Training}
Preparing pretraining documents with quality annotations is the first key step in building a quality classifier~\citep{dubey2024llama,abdin2024phi,yang2024qwen2}. Similar to the work ~\citep{penedo2024fineweb}\footnote{We use the same 460K document samples as in the FineWeb-Edu-Annotation dataset.}, we constructed two versions of quality annotation data. We prompt Mistral 8x22B-instruct\footnote{\url{https://mistral.ai/news/mixtral-8x22b/}} and Nemotron-340B-instruct~\citep{adler2024nemotron}, to score web documents from FineWeb based on their educational value on a scale from 0 to 5.  We then fine-tune a linear regression model on top of the Snowflake-arctic-embed-m embedding model~\citep{merrick2024arctic} using the two different version of training sets. The two models have been trained for 20 epochs with a learning rate of 3e-4, with the embedding and encoder layers frozen, and we selected the checkpoint with the highest F1 score on
the held-out validation set.

 We also employ the DCLM classifier which is a fastText-based classifier released by~\citet{li2024datacomp}. The DCLM classifier is trained on a combination of instruction-formatted data~\citep{OpenHermes} and high-scoring posts data from ELI5 subreddit~\citep{fan2019eli5}, and has shown stronger performance in identifying high-quality pretraining tokens, compared to the FineWeb-Edu classifier ~\citep{penedo2024fineweb}. The DCLM classifier will offer a new perspective in labeling high-quality pretraining documents, and will help increase the recall of high-quality tokens.

\paragraph{Quality Scoring and Bucketing}
First, we use each of the three classifiers to predict the quality scores for all the documents. Then based on the ranked quality score from each classifier, we rounded the model's output score to integers from 0 to 19. So that each score bucket will have around 5\% of the documents, and bucket 19 will have the top 5\% highest quality documents. We then assign the final quality score for each document by ensembling the three classifiers' integer score by a maximum operation. The number of documents distribution in each buckets will be skewed by the ensemble operation.

\paragraph{Quality Labeling}
In order to assign a quality label that is more aligned with their real performance on downstream tasks, we further group the fine-grained quality score predicted by three classifiers into 5 downstream quality categories. We used annealing to assess each data bucket's downstream task's quality. Specifically, we measure the quality of each bucket by continuous pretraining with 50B tokens on a 70\% trained 8B models. We assign 66\% of weight to the default data mix and 34\% to the dataset that we are evaluating. By comparing the average performance of each bucket over 9 tasks, we group the 20 buckets into 5 big categories, with the final distribution shown in Table~\ref{table:quality_label_stats}. For more details, please see Appendix~\ref{appendix:bucketcomparison}.

\begin{table}[htbp]\centering\small
\begin{tabularx}{\linewidth}{Xcrr}
\toprule
\textbf{Quality Label} & \multicolumn{1}{l}{\textbf{Buckets}} & \multicolumn{1}{l}{\textbf{\# Tokens (B)}} & \multicolumn{1}{l}{\textbf{Token ($\%$)}} \\ \midrule
High                   & 19                                   & 553                                        & 12.63                                     \\
Medium-High            & 18                                   & 504                                        & 11.52                                     \\
Medium                 & 12-17                                & 2,023                                      & 46.24                                     \\
Medium-Low             & 7-11                                 & 894                                        & 20.43                                     \\
Low                    & 0-6                                  & 402                                        & 9.18                                      \\ \bottomrule
\end{tabularx}
\caption{Common Crawl quality labels statistics.}
\label{table:quality_label_stats}
\end{table}

\subsection{Synthetic Data Generation} \label{sec:synthetic}

Upon reviewing samples across the quality tiers, we observe that documents with lower scores tend to contain more noise and errors, while those scoring higher generally exhibit good writing and formatting.
Therefore, we employ different strategies when generating data from low- and high-quality documents.

For low-quality data, our goal is to improve the quality by reducing noise and errors while preserving useful information, thereby decreasing training compute expenses. As shown by~\citet{maini2024rephrasing}, rephrasing web data using a medium-sized language model yields an enhanced parallel corpus of synthetic data, thereby reducing model perplexity and boosting its accuracy on downstream tasks.
Unlike existing methods that create new content such as textbooks and short stories~\cite{wang2023self,eldan2023tinystories,gunasekar2023textbooks}, our rephrasing-based approach does not utilize the language model as a knowledge bank but focuses on transforming provided texts into another style, allowing it to operate with a lighter-weight model.
We adopt the \textrm{Wikipedia} style prompt from~\citep{maini2024rephrasing}
to rewrite low-quality documents (Prompt~5 in Appendix~\ref{sec:appendix-prompt-templates}), which effectively reduces errors and redundancies and improves formatting.

For high-quality data, we aim to obtain more unique tokens and condense essential knowledge. According to~\citep{muennighoff2024scaling}, adding repeated tokens yields a diminishing return, especially after 4 epochs.
For high-quality documents, we generate synthetic data using four additional prompts:
(1) \textrm{Diverse Question-Answer (QA) pairs}: ask questions in various forms (e.g., yes/no question, open-ended question, multi-choice question) about factual information in the text and provide the correct answers;
(2) \textrm{Distill}: rewrite the text into a concise and clear passage;
(3) \textrm{Extract knowledge}: rewrite knowledge from the text and disregard uninformative content;
(4) \textrm{Knowledge list}: extract key information from the text as an organized list.
We require the model to provide clear and concise responses while preserving factual information and concrete details such as numbers.
The full prompts are shown in Appendix~\ref{sec:appendix-prompt-templates}.

As we increase the length of provided text, the model shows a tendency to produce over-simplified outputs with reduced detail. Therefore, we chunk each document into segments, each of which contains one or more complete lines and is shorter than a specific token limit.\footnote{The token limit is set to 512 for \textrm{Wikipedia}, 2,000 for \textrm{Distill}, 1,400 for \textrm{Extract Knowledge} and 1,000 for \textrm{Diverse QA Pairs} and \textrm{Knowledge List}, including tokens from the prompt and chat format.} Over-length lines exceeding the token limit are discarded.

\begin{tcolorbox}[boxsep=0mm,left=1mm,right=1mm,top=1mm,bottom=1mm,fontupper=\small\ttfamily,halign=left,arc=0mm,boxrule=1pt,colback=gray!10]
\textbf{Question}: Which year did the United Nations implement the 2030 agenda for SDGs?

\textbf{Answer}: January 1, 2016

\vspace{4pt}

\textbf{Question}: What are the three key dimensions of sustainable development covered by the SDGs?

\textbf{Answer}: (a) economic growth, (b) social inclusion, and (c) environmental protection

\vspace{4pt}

\textbf{Question}: Which of the following can flossing prevent? A) Cavities B) Gum disease C) Both A and B D) Neither A nor B

\textbf{Answer}: C) Both A and B

\vspace{4pt}

\textbf{Question}: Is flossing important even if you brush your teeth twice a day?

\textbf{Answer}: Yes, flossing is important as it reaches areas that brushing alone cannot.
\end{tcolorbox}
\noindent\begin{minipage}{\linewidth}
\captionof{figure}{Examples of generated question-answer pairs.}
\end{minipage}

Our post-processing steps include removing incomplete results, eliminating specific Markdown formatting (e.g., double asterisks), stripping away prefixes of certain patterns (e.g., ``\textit{Here is a paraphrased version:}'' and ``\textit{Paraphrased Text:}''), removing quotation marks enclosing the entire response, and filtering out under-length outputs (i.e., shorter than 50 tokens).
For \textrm{Wikipedia} results, we concatenate passages generated from segments belonging to the same original document.
For \textrm{Diverse QA Pairs} results, we shuffle the generated question and answer pairs, retain up to a number based on the length of the segment, and append the pairs to the end of the segment.

Using the instruct version of Mistral NeMo 12B\footnote{\url{https://mistral.ai/news/mistral-nemo}} with FP8 inference, a top-p value of $0.9$, and a sampling temperature of $0.5$, we synthesize over 1.8T tokens as Table~\ref{table:synthetic_data_stats} shows, including 336.3B tokens from low-quality documents and 1.5T tokens from high-quality documents.  We do not use medium-quality documents for synthetic data generation due to time and resource constraints. We employ TensorRT-LLM\footnote{\url{https://github.com/NVIDIA/TensorRT-LLM}} and NeMo-Skills\footnote{\url{https://github.com/NVIDIA/NeMo-Skills}} to enable large-scale data synthesis.

\begin{table}[!hbt]
    \small
    \centering
    \begin{tabularx}{\linewidth} {l c X c}
    \toprule
    \textbf{Source}      & \textbf{\#Raw}          & \textbf{Prompt}            & \textbf{\#Synthetic} \\
    \midrule
    \textrm{Low}  & 403.0          & \textrm{Wikipedia}      & 336.3 \\
    \midrule
    \multirow{5}{*}{\textrm{High}}  &  \multirow{5}{*}{451.3} & \textrm{Wikipedia}         & 372.9 \\
                         &                         & \textrm{Diverse QA Pairs}  & 499.5 \\
                         &                         & \textrm{Distill}           & 157.6 \\
                         &                         & \textrm{Extract Knowledge} & 303.6 \\
                         &                         & \textrm{Knowledge List}    & 203.2 \\
    \bottomrule
    \end{tabularx}
    \caption{Synthetic data token count statistics (billion).}
    \label{table:synthetic_data_stats}
\end{table}

\subsection{Putting It All Together}

\begin{table}[hbt]\small\centering
\begin{tabularx}{\linewidth}{Xccc}
\toprule
\textbf{Dataset} & \textbf{Total} & \textbf{Unique} & \textbf{Synthetic} \\ \midrule
FineWebEdu-2  & 5.4  & 1.1   & -         \\
FineWebEdu    & 1.3  & 0.2   & -         \\
\DCLM & 3.8  & 1.0     & -         \\
\Ours          & 6.3  & 4.4   & 1.9      \\
\Ours-HQ       & 1.1  & 0.6   & 0.5      \\ \bottomrule
\end{tabularx}
\caption{Dataset sizes in trillions of tokens. "Unique" shows the estimated number of tokens after global fuzzy deduplication of the real tokens.}
\label{table:dataset_sizes}
\end{table}

Combining the techniques above to the 99 snapshots CC-MAIN-2013-20 through CC-MAIN-2024-30 of Common Crawl, we create a 6.3T token dataset (\Ours), consisting of 4.4T globally deduplicated tokens and 1.9T synthetically derived tokens. This dataset has roughly 4$\times$ more unique tokens than FineWebEdu-2 and \DCLM, since both of those datasets only underwent a sharded form of approximate deduplication and contain roughly $80\%$ fuzzy duplicates~\citep{fineweb-edu-discussion,li2024datacomp}. To enable a fairer comparison over relatively short token horizons, we thus also consider a 1.1T token high quality subset of our data (\Ours-HQ), consisting of just the highest-scoring real and diverse QA pairs synthetic data. The size breakdown of the datasets is shown in Table~\ref{table:dataset_sizes}.

\section{Experiments}

\begin{table*}[!hbt]\small\centering
\setlength{\tabcolsep}{5pt}
\begin{tabularx}{\textwidth}{Xlllllllllll}
\toprule
\textbf{Dataset}        & \textbf{ARC-E} & \textbf{ARC-C} & \textbf{H}    & \textbf{W}    & \textbf{RACE} & \textbf{PIQA} & \textbf{SIQA} & \textbf{CSQA} & \textbf{OBQA} & \textbf{MMLU} & \textbf{Avg}       \\ \midrule
FineWebEdu-2          & 71.9          & 44.7          & 75.4          & 67.0          & 36.8          & 79.5          & 45.2          & 25.5           & 43.8          & 42.4          & 53.2          \\
FineWebEdu            & 73.6          & 48.0          & 70.7          & 64.6          & \textbf{38.0} & 76.4          & 43.5          & 30.0           & 44.4          & 42.9          & 53.2          \\
\DCLM                 & 74.7          & 47.0          & 76.3          & 69.1          & 36.5          & 79.7          & 45.6          & 44.1           & 44.0          & 53.4          & 57.0          \\
\Ours                 & 75.3          & 50.7          & 75.9          & 67.8          & 37.9          & \textbf{80.5} & 45.1          & 47.7           & 44.2          & 53.0          & 57.8          \\
\Ours-HQ              & \textbf{78.8} & \textbf{52.9} & \textbf{76.6} & \textbf{69.4} & 36.4          & 80.1          & \textbf{46.6} & \textbf{55.8}  & \textbf{45.4} & \textbf{59.0} & \textbf{60.1} \\ \bottomrule
\end{tabularx}
\caption{Results for 8B parameter models trained on 1T tokens ($73\%$ English Common Crawl from the tested dataset, $27\%$ the same, fixed non-Crawl datasets). The models were evaluated on ARC-Easy, ARC-Challenge, Hellaswag, Winogrande, RACE, PIQA, Social IQA, Commonsense QA, Openbook QA, and MMLU.}
\label{table:main_detailed}
\end{table*}

\begin{table*}[!hbt]\small\centering
\begin{tabularx}{\textwidth}{Xlllllllllll}
\toprule
\textbf{Model}        & \textbf{ARC-E} & \textbf{ARC-C} & \textbf{H}    & \textbf{W}    & \textbf{RACE} & \textbf{PIQA} & \textbf{SIQA} & \textbf{CSQA} & \textbf{OBQA} & \textbf{MMLU} & \textbf{Avg}       \\ \midrule
Llama 3.1 & 82.4          & 55.0          & 79.3          & \textbf{74.7} & \textbf{39.1} & \textbf{81.2} & \textbf{48.3} & \textbf{70.6} & \textbf{46.0} & 65.3          & 64.2          \\
Ours       & \textbf{82.7} & \textbf{58.1} & \textbf{80.8} & 73.8          & 37.8          & 81.1          & 47.4          & 69.9          & 45.4          & \textbf{70.3} & \textbf{64.7} \\ \bottomrule
\end{tabularx}
\caption{Comparison of our 8B parameter model vs Llama 3.1 8B. Both were trained for 15T tokens. The numbers for Llama 3.1 are from our own lm-evaluation-harness setup described in Section~\ref{section:experiment-setup} and may not match Meta's publicly reported numbers, as Meta made various customizations to the benchmarks.}
\label{table:8b-15t}
\end{table*}

\subsection{Experiment Setup}
\label{section:experiment-setup}

\paragraph{Training Setup} We use the open source Megatron-LM library\footnote{\url{https://github.com/NVIDIA/Megatron-LM}}~\citep{shoeybi2019megatron} to train standard 8B parameter transformer LLMs. The hyperparameter details are shown in Appendix~\ref{appendix:hyperparameters}.

\paragraph{Data Blend} Unless otherwise noted, we train for 1T tokens on a blend of $73\%$ English Common Crawl data and $27\%$ a fixed mix of specialized code, papers, books, patents, and Wikipedia datasets~\citep{adler2024nemotron}. When comparing datasets, we vary only the $73\%$ English Common Crawl portion. See Table~\ref{table:datablend} in Appendix~\ref{appendix:hyperparameters}.

\paragraph{Evaluation Setup} We use the open source LM Evaluation Harness library\footnote{\url{https://github.com/EleutherAI/lm-evaluation-harness}}~\citep{eval-harness} to evaluate on the following ten common sense and reasoning tasks (reported metric in parentheses): ARC-Easy and ARC-Challenge (normalized accuracy)~\citep{clark2018think}, Hellaswag (normalized accuracy)~\citep{zellers2019hellaswag}, Winogrande (accuracy)~\citep{sakaguchi2021winogrande}, RACE (accuracy)~\citep{lai2017race}, PIQA (normalized accuracy)~\citep{bisk2020piqa}, Social IQA (accuracy)~\citep{sap2019social}, Commonsense QA (accuracy)~\citep{talmor2019commonsenseqa}, Openbook QA (normalized accuracy)~\citep{mihaylov2018can}, and MMLU (accuracy)~\citep{hendrycks2021measuring}.

\subsection{Main Results}

\paragraph{Short Token Horizon (1T)}

To validate the quality of our datasets, we first train standard 8B parameter transformer LLMs over a relatively short 1T token horizon. The results are shown in Table~\ref{table:main_detailed}. Our high quality dataset (\Ours-HQ) shows accuracy gains over \DCLM and FineWeb-Edu on all tasks except RACE. In particular, there is a 5.6 MMLU and 3.1 average gain over \DCLM. This shows the effectiveness of our classifier ensembling and synthetic data even in the non-data-constrained setting. Our complete 6.3T token dataset (\Ours) gives MMLU and average accuracies roughly on par with \DCLM. But since this dataset contains 4$\times$ more unique real tokens, we expect it to be superior in data-constrained settings like 15T token training runs.

\paragraph{Long Token Horizon (15T)}

Our dataset contributed 7.2T of the tokens used to train an 8B model for 15T tokens.
As shown in Table~\ref{table:8b-15t}, our model achieves a higher average accuracy than Llama 3.1 8B, which was also trained for 15T tokens, including an MMLU score of 70.3 vs. Llama's 65.3. This shows that our dataset is indeed suitable for state-of-the-art training over long token horizons. For more details on this experiment, please see Appendix~\ref{appendix:8b15t}.

\subsection{Ablation Study} \label{sec:ablation}
To further investigate the contribution and effect of each module in our method, we conducted thorough ablation studies.

\paragraph{Extractor \& Filter Comparison}

As we have discussed in Section~\ref{sec:extract_filter}, by deploying Justext instead of Trafilatura and removing filter from the post-processing step, we can attain significantly 57.4\% more high-quality tokens. We also conduct ablation studies to better understand the impact of the extractor selection and the removal of filter through downstream benchmarks. We carry out four 8B-1T experiments. We report the benchmark scores in Table~\ref{table:extactor_filter_ablation}. Beyond the token-yield benefit by leveraging Justext instead of Trafilatura and not using heuristic filters, we see that combining these two does not impact the downstream task accuracies with only marginal differences (comparing Trafilatura filtered vs. Justext unfiltered). Moreover, when we ONLY remove filter from high-quality tokens, the results get further improved (comparing Justext unfiltered vs. Justext HQ unfiltered). In particular, MMLU gets boosted by +2\%. Note that, the motivation behind removing filter is to boost token yield, especially on high-quality tokens due to the notable scarcity of such. Given the experimental results and considering the overall growth in token yield, we opt to only remove filter from high-quality tokens.

\begin{table}[!hbt] \small \centering
\begin{tabularx}{\linewidth}{Xcc}
\toprule
\textbf{Exp name}     & \textbf{MMLU} & \textbf{Avg (non-MMLU)} \\ \midrule
Trafilatura filtered  & 55.4              & 60.6                   \\ 
Justext filtered      & 54.1              & \textbf{60.9}          \\ 
Justext unfiltered    & 55.5              & 60.3                   \\ 
Justext HQ-unfiltered & \textbf{57.5}     & 60.6                   \\ \bottomrule
\end{tabularx}
\caption{Ablation studies on extractor and filter. HQ means high-quality data judged by FineWeb-Edu classifier (score 3,4,5). HQ-unfiltered means filtering is applied only to LQ data. See Appendix \ref{sec:non_mmlu} for more details.}
\label{table:extactor_filter_ablation}
\end{table}

\paragraph{Classifiers Comparison}
Assembling different classifiers to label the document quality is one of the key steps in constructing our datasets, so we did thorough analysis and comparison of the component.

We did a detailed comparison of two types of classifiers that we employ in our method: the FineWeb-Edu classifier which score document quality based on their educational-level, and the DCLM-based classifier which value the informativeness of the document.  We compare the high-quality documents predicted by the two classifiers on a randomly selected Common Crawl Snapshot (CC-MAIN-2021-21). Table~\ref{table:hq-doc-comparision} shows the document statistics comparison. We can see that only 10\% of the documents are predicted as high quality by both classifiers, while 35.4\% documents are predicted as high quality by FineWeb-Edu classifier only, and 54.4\% of documents are predicted as high-quality by DCLM classifier. Therefore, ensembling different classifiers can increase the recall of  high-quality documents from Common Crawl.\footnote{Detailed URL domain comparison can be found in Appendix~\ref{sec:appendix-1}}

We further compare each of the classifiers with the ensembled method\footnote{Note that we did not employ FineWeb-Edu classifier in our ensemble for license issue, since it is trained with annotations from Llama3.} by their downstream tasks' performances.  We pretrain 8B parameters LLMs with 1T tokens, using the high-quality documents labeled by different classifiers on randomly selected 13 Common Crawl snapshots (see Appendix~\ref{sec:13_snapshots}). Table~\ref{table:classifier_ablation} shows the detailed comparison on different evaluation tasks. We can see that the ensembled method greatly boost the high-quality tokens percentage from 9\% to 25\%, while still achieving the highest general language understanding performance on average on all the tasks. The ensembled method also outperforms the FineWeb-Edu classifier and the DCLM classifier, in terms of the high-quality token percentage, and is on-par or slightly better on the 9 evaluation tasks. This is very important since more unique high-quality tokens is the key in pretraining larger LLMs on longer tokens horizons. 

\begin{table}[!hbt]
\small
\centering
\begin{tabularx}{\linewidth}{Xcc}
\toprule
\textbf{What} & \multicolumn{1}{l}{\textbf{\#Docs}} & \multicolumn{1}{l@{}}{\textbf{Total unique(\%)}} \\ \midrule
Total unique in union & 11,359,655 & 100.0\% \\
In intersection & 1,152,821 & 10.1\% \\
In FineWeb-Edu only & 4,022,294 & 35.4\% \\
In DCLM only & 6,184,540 & 54.4\% \\ \bottomrule
\end{tabularx}
\caption{High-quality documents overlap analysis.}
\label{table:hq-doc-comparision}
\end{table}

\addtolength{\tabcolsep}{-3pt}   
\begin{table*}[!ht]
\centering
\small
\begin{tabularx}{\textwidth}{Xcccccccccccc}
\toprule
\textbf{Classifier} & \textbf{HQ(\%)} & \textbf{ARC-E} & \textbf{ARC-C} & \textbf{H} & \textbf{W} & \textbf{RACE} & \textbf{PIQA} & \textbf{SIQA} & \textbf{CSQA} & \textbf{OBQA} & \textbf{MMLU} & \textbf{Avg} \\ \midrule
FineWeb-Edu & 8\% & 77.7 & 50.1 & 74.9 & 67.3 & \textbf{39.5} & 78.8 & 45.8 & 53.6 & 43.0 & 55.4 & 59.0 \\
DCLM & 11\% & 76.0 & 49.2 & \textbf{76.5} & \textbf{70.2} & 38.2 & \textbf{80.8} & 33.9 & 55.2 & 45.8 & 56.0 & 58.4 \\
Ours-mistral & 9\% & 75.8 & 49.2 & 75.9 & 66.9 & 37.5 & 80.1 & \textbf{46.2} & 46.9 & 44.8 & 53.2 & 58.1 \\
Ours-nemotron-340B & 14\% & 76.3 & \textbf{50.3} & 75.6 & 67.5 & 37.8 & 80.2 & 34.3 & 54.0 & \textbf{46.2} & 54.9 & 58.0 \\
Ours-ensembled & \textbf{25\%} & \textbf{78.0} & 49.7 & 75.3 & 67.1 & 37.2 & 79.6 & 45.7 & \textbf{56.8} & 44.8 & \textbf{56.4} & \textbf{59.4} \\ \bottomrule
\end{tabularx}
\caption{Different classifiers comparison. Our ensemble method includes the three classifiers: Ours-mistral, Ours-nemotron-340B and DCLM.}
\label{table:classifier_ablation}
\end{table*}
\addtolength{\tabcolsep}{3pt}

\begin{table*}[!hbt]
\small
\centering
\begin{tabularx}{\textwidth}{Xccccccccccc}
\toprule
\textbf{Blend}              & \textbf{ARC-E}         & \textbf{ARC-C}         & \textbf{H}     & \textbf{W}    & \textbf{RACE}          & \textbf{PIQA}          & \textbf{SIQA}    & \textbf{CSQA} & \textbf{OBQA}    & \textbf{MMLU}          & \textbf{Avg}       \\

\midrule
LQ-Base & 67.7 & 41.8 & \textbf{75.2} & \textbf{67.1} & 37.4 & 78.8 & 45.3 & 36.9 & 41.0 & \textbf{48.2} & 52.5 \\
LQ-Synthetic & \textbf{71.3} & \textbf{45.2} & 75.0 & 66.9 & 37.4 & \textbf{79.4} & \textbf{46.2} & \textbf{41.6} & \textbf{42.8} & 47.1 & \textbf{54.0} \\
\midrule
HQ-Base  & 74.2 & 47.7 & \textbf{74.8} & 66.9 & 37.3 & 78.2 & \textbf{46.0} & 47.3 & 43.6 & 53.4 & 55.8 \\
HQ-Synthetic  & \textbf{76.7} & \textbf{49.2} & 74.5 & \textbf{67.3} & \textbf{38.2} & \textbf{78.8} & 45.2 & \textbf{47.9} & \textbf{45.8} & \textbf{53.6} & \textbf{56.7} \\
\bottomrule
\end{tabularx}
\caption{Impact of incorporating synthetic data.}
\label{table:synthetic_data_ablation}
\end{table*}

\paragraph{Evaluating Synthetic Data} 
As Table~\ref{table:synthetic_data_ablation} shows, this ablation study aim to answer two questions:
(1) Does rephrasing low-quality improve accuracies on downstream tasks?
(2) Can synthetic data help offset the decreasing value of duplicated data reported in~\citep{muennighoff2024scaling}?
To answer these questions, we train four 8B models with the same hyperparameters on different blends of 1T tokens:
(1) LQ-Base: original Common Crawl data including low-quality documents;
(2) LQ-Synthetic: an augmented version of LQ-Base where the low-quality documents are rephrased;
(3) HQ-Base: a blend containing eightfold high-quality documents and less low- and medium-quality documents;
(4) HQ-Synthetic: a variant of HQ-Base where 4 repetitions of the high-quality documents are swapped out for synthetic datasets.

By comparing the results between LQ-Base and LQ-Synthetic, we can see that rephrasing low-quality data leads to 1.50 absolute gains on average score. We also observe noticeable boosts from 1.80\% to 4.75\% on ARC-Easy, ARC-Challenge, OpenbookQA, CommonsenseQA; however, we also encounter slight accuracy drops on some tasks, which may indicate potential misinformation introduced by data synthesis.
Current practices typically utilize data curation approaches to detect and eliminate noisy examples. Due to time and resource constraints, we leave the detailed exploration of this issue for future efforts.

The comparison between HQ-Base and HQ-Synthetic shows that swapping 4 out of 8 epochs of high-quality data with a mix of synthetic datasets improves accuracy on most benchmarks.
This improvement could potentially result from two factors: the incorporation  of fresh unique tokens and styles that enable the model to learn specific abilities (e.g., question answering) or absorb knowledge more efficiently.

\section{Related Work}
The Phi series of models pioneered training on small amounts of very high quality data, including curated Web and synthetic data~\citep{gunasekar2023textbooks,li2023textbooks,abdin2024phi}. However, their focus is on shorter token horizons and they share limited details. FineWeb-Edu and \DCLM are the main points of comparison for our paper~\citep{li2024datacomp,penedo2024fineweb}. We build upon their core idea of model-based filtering, but show how to improve the filtering and data quantity through a combination of other techniques. Other English Common Crawl datasets such as C4, DOLMA, Gopher, RefinedWeb, TxT360 largely focus on extraction and non-learned heuristics~\citep{penedo2023refinedweb,soldaini2024dolma,rae2021scaling,raffel2020exploring,txt360data2024}. Just as for FineWeb-Edu and \DCLM, the core pipeline we started from incorporates many of these ideas, but our paper describes how to modify and go beyond these non-learned techniques to achieve state-of-the-art accuracy and diversity. Concurrent work Zyda-2 shows how to filter, cross-deduplicate, and combine the FineWeb-Edu, \DCLM, Zyda-1, and Dolma-CC datasets into a higher-accuracy and larger whole~\citep{tokpanov2024zyda25trilliontoken}. In contrast, we focus on techniques for the creation of a new English Common Crawl dataset rather than combinations or modifications of existing datasets. Finally, many works have focused on creating multilingual datasets~\citep{xue-etal-2021-mt5,brack-etal-2024-community,abadji-etal-2022-towards,wenzek-etal-2020-ccnet,kudugunta2023madlad400multilingualdocumentlevellarge}. We leave extension of our ideas beyond English to the future.

Synthetic datasets have been widely used in language model pre-training and post-training. In~\citep{cheng2024instructionpretraininglanguagemodels}, instruction-response pairs are synthesized for pre-training.
In~\citep{eldan2023tinystories}, the authors show that smaller or simpler models trained on a synthetic dataset of short stories are capable of generating fluent and consistent stories. Similarly, smaller models trained using high-quality synthetic textbook and exercise datasets can achieve impressive high accuracy on coding benchmarks~\citep{gunasekar2023textbooks,li2023textbooks}.
These approaches typically require a powerful language model, such as GPT-3.5 and GPT-4 in~\citep{eldan2023tinystories}, to synthesize new contents.
Instead, \citep{maini2024rephrasing} shows that compact models such as Qwen-1.8B and Mistral-7B are adequate to rephrase web data. This approach generates diverse, high-quality synthetic data that effectively lowers model perplexity and boosts performance across benchmarks.
We adopt this main idea, but explore more prompts and show how to specialize them for low and high quality data.

\section{Conclusion}

For producing long-horizon pretraining tokens for LLMs from English Common Crawl data, we showed how to improve upon the state of the art and achieve better trade-offs between benchmark accuracy and data quantity, as measured by number of unique real tokens. Specifically, we showed the efficacy of ensembling model-based quality filters, rephrasing low and high quality documents, and reducing the reliance on non-learned heuristics. The dataset is public and split by quality level and type (actual data vs. different types of synthetic data), enabling the community to do further experiments on quality vs. diversity and how to build effective short and long horizon curricula.

\section{Limitations}

Some of the key limitations of our work are as follows. For the model-based filter ensembling and quality bucketing, we only had time and resources to try a single strategy. Though it is effective, it is possible this could be improved upon in future work, especially to improve the sensitivity at the higher-quality end of the spectrum. For the rephrased data, we did not verify the factual accuracy or fidelity to the original contents. More work is required to understand the risks of hallucinations or loss of content diversity in this setting and how to mitigate them. We also only looked at rephrasing low and high quality data. It could be interesting to explore how to best rephrase medium quality data as well. We did not do ablations on all parts of the pipeline. There is probably room for improvement with, for example, the language identification. Overall, we tried our methods only on English text. More work is needed to adapt our methods to other languages.

Finally, we did not decontaminate the dataset, as there is not yet a strong consensus on how to best do this and the impact is uncertain and debated, especially for large models trained over large token horizons. We note that the datasets we compare against (FineWeb-Edu, DCLM) were released without decontamination, and the model we compare against (Meta Llama 3.1) was also trained on contaminated data. DCLM reports some contamination analysis, but the findings suggest contamination is not a key factor: e.g., MMLU actually increases after decontamination, and DCLM does better than FineWeb on MMLU, even though FineWeb has more MMLU contamination (see Section 4.6 and Appendix N in~\citet{li2024datacomp}). Still, it would be interesting to better understand the impact of contamination for different model sizes and different token horizons, and we hope the community can explore such questions on this public dataset.

\section*{Acknowledgments}
We thank the Common Crawl Foundation for hosting the dataset. We thank Pedro Ortiz Suarez for valuable feedback that improved the paper and Greg Lindahl for help with improving the data formatting and layout. Furthermore, we thank Ayush Dattagupta, Randy Gelhausen, Smita Ithape, Vibhu Jawa, Nirmal Kumar Juluru, Vineeth Kalluru, Mehran Maghoumi, Arham Mehta, Ranjit Rajan, Janaki Vamaraju, Ryan Wolf, and Sarah Yurick for help with NeMo Curator and for open sourcing Nemotron-CC recipes and classifiers as part of NeMo Curator.

\bibliography{custom}

\begin{thebibliography}{50}
\providecommand{\natexlab}[1]{#1}

\bibitem[{Abadji et~al.(2022)Abadji, Ortiz~Suarez, Romary, and Sagot}]{abadji-etal-2022-towards}
Julien Abadji, Pedro Ortiz~Suarez, Laurent Romary, and Beno{\^\i}t Sagot. 2022.
\newblock \href {https://aclanthology.org/2022.lrec-1.463} {Towards a cleaner document-oriented multilingual crawled corpus}.
\newblock In \emph{Proceedings of the Thirteenth Language Resources and Evaluation Conference}, pages 4344--4355, Marseille, France. European Language Resources Association.

\bibitem[{Abdin et~al.(2024)Abdin, Jacobs, Awan, Aneja, Awadallah, Awadalla, Bach, Bahree, Bakhtiari, Behl et~al.}]{abdin2024phi}
Marah Abdin, Sam~Ade Jacobs, Ammar~Ahmad Awan, Jyoti Aneja, Ahmed Awadallah, Hany Awadalla, Nguyen Bach, Amit Bahree, Arash Bakhtiari, Harkirat Behl, et~al. 2024.
\newblock Phi-3 technical report: A highly capable language model locally on your phone.
\newblock \emph{arXiv preprint arXiv:2404.14219}.

\bibitem[{Adler et~al.(2024)Adler, Agarwal, Aithal, Anh, Bhattacharya, Brundyn, Casper, Catanzaro, Clay, Cohen et~al.}]{adler2024nemotron}
Bo~Adler, Niket Agarwal, Ashwath Aithal, Dong~H Anh, Pallab Bhattacharya, Annika Brundyn, Jared Casper, Bryan Catanzaro, Sharon Clay, Jonathan Cohen, et~al. 2024.
\newblock Nemotron-4 340b technical report.
\newblock \emph{arXiv preprint arXiv:2406.11704}.

\bibitem[{Ainslie et~al.(2023)Ainslie, Lee-Thorp, de~Jong, Zemlyanskiy, Lebron, and Sanghai}]{ainslie2023gqa}
Joshua Ainslie, James Lee-Thorp, Michiel de~Jong, Yury Zemlyanskiy, Federico Lebron, and Sumit Sanghai. 2023.
\newblock Gqa: Training generalized multi-query transformer models from multi-head checkpoints.
\newblock In \emph{Proceedings of the 2023 Conference on Empirical Methods in Natural Language Processing}, pages 4895--4901.

\bibitem[{Barbaresi(2021)}]{barbaresi-2021-trafilatura}
Adrien Barbaresi. 2021.
\newblock \href {https://aclanthology.org/2021.acl-demo.15} {{Trafilatura: A Web Scraping Library and Command-Line Tool for Text Discovery and Extraction}}.
\newblock In \emph{Proceedings of the Joint Conference of the 59th Annual Meeting of the Association for Computational Linguistics and the 11th International Joint Conference on Natural Language Processing: System Demonstrations}, pages 122--131. Association for Computational Linguistics.

\bibitem[{{Ben Allal}(2024)}]{fineweb-edu-discussion}
Loubna {Ben Allal}. 2024.
\newblock Most of the data is duplicated?
\newblock \url{https://huggingface.co/datasets/HuggingFaceFW/fineweb-edu/discussions/7}.
\newblock Accessed: October 24, 2024.

\bibitem[{Bisk et~al.(2020)Bisk, Zellers, Gao, Choi et~al.}]{bisk2020piqa}
Yonatan Bisk, Rowan Zellers, Jianfeng Gao, Yejin Choi, et~al. 2020.
\newblock Piqa: Reasoning about physical commonsense in natural language.
\newblock In \emph{Proceedings of the AAAI conference on artificial intelligence}, volume~34, pages 7432--7439.

\bibitem[{Brack et~al.(2024)Brack, Ostendorff, Ortiz~Suarez, Saiz, Castilla, Palomar-Giner, Shvets, Schramowski, Rehm, Villegas, and Kersting}]{brack-etal-2024-community}
Manuel Brack, Malte Ostendorff, Pedro Ortiz~Suarez, Jos{\'e}~Javier Saiz, I{\~n}aki~Lacunza Castilla, Jorge Palomar-Giner, Alexander Shvets, Patrick Schramowski, Georg Rehm, Marta Villegas, and Kristian Kersting. 2024.
\newblock \href {https://aclanthology.org/2024.mrl-1.19} {Community {OSCAR}: A community effort for multilingual web data}.
\newblock In \emph{Proceedings of the Fourth Workshop on Multilingual Representation Learning (MRL 2024)}, pages 232--235, Miami, Florida, USA. Association for Computational Linguistics.

\bibitem[{Cheng et~al.(2024)Cheng, Gu, Huang, Bi, Huang, and Wei}]{cheng2024instructionpretraininglanguagemodels}
Daixuan Cheng, Yuxian Gu, Shaohan Huang, Junyu Bi, Minlie Huang, and Furu Wei. 2024.
\newblock \href {https://doi.org/10.18653/v1/2024.emnlp-main.148} {Instruction pre-training: Language models are supervised multitask learners}.
\newblock In \emph{Proceedings of the 2024 Conference on Empirical Methods in Natural Language Processing}, pages 2529--2550, Miami, Florida, USA. Association for Computational Linguistics.

\bibitem[{Clark et~al.(2018)Clark, Cowhey, Etzioni, Khot, Sabharwal, Schoenick, and Tafjord}]{clark2018think}
Peter Clark, Isaac Cowhey, Oren Etzioni, Tushar Khot, Ashish Sabharwal, Carissa Schoenick, and Oyvind Tafjord. 2018.
\newblock Think you have solved question answering? try arc, the ai2 reasoning challenge.
\newblock \emph{arXiv preprint arXiv:1803.05457}.

\bibitem[{Dubey et~al.(2024)Dubey, Jauhri, Pandey, Kadian, Al-Dahle, Letman, Mathur, Schelten, Yang, Fan et~al.}]{dubey2024llama}
Abhimanyu Dubey, Abhinav Jauhri, Abhinav Pandey, Abhishek Kadian, Ahmad Al-Dahle, Aiesha Letman, Akhil Mathur, Alan Schelten, Amy Yang, Angela Fan, et~al. 2024.
\newblock The llama 3 herd of models.
\newblock \emph{arXiv preprint arXiv:2407.21783}.

\bibitem[{Eldan and Li(2023)}]{eldan2023tinystories}
Ronen Eldan and Yuanzhi Li. 2023.
\newblock Tinystories: How small can language models be and still speak coherent english?
\newblock \emph{arXiv preprint arXiv:2305.07759}.

\bibitem[{Fan et~al.(2019)Fan, Jernite, Perez, Grangier, Weston, and Auli}]{fan2019eli5}
Angela Fan, Yacine Jernite, Ethan Perez, David Grangier, Jason Weston, and Michael Auli. 2019.
\newblock Eli5: Long form question answering.
\newblock In \emph{Proceedings of the 57th Annual Meeting of the Association for Computational Linguistics}, pages 3558--3567.

\bibitem[{Feng et~al.(2024)Feng, Prabhumoye, Kong, Su, Patwary, Shoeybi, and Catanzaro}]{feng2024maximizedataspotentialenhancing}
Steven Feng, Shrimai Prabhumoye, Kezhi Kong, Dan Su, Mostofa Patwary, Mohammad Shoeybi, and Bryan Catanzaro. 2024.
\newblock \href {https://arxiv.org/abs/2412.15285} {Maximize your data's potential: Enhancing llm accuracy with two-phase pretraining}.
\newblock \emph{Preprint}, arXiv:2412.15285.

\bibitem[{Gao et~al.(2023)Gao, Tow, Abbasi, Biderman, Black, DiPofi, Foster, Golding, Hsu, Le~Noac'h, Li, McDonell, Muennighoff, Ociepa, Phang, Reynolds, Schoelkopf, Skowron, Sutawika, Tang, Thite, Wang, Wang, and Zou}]{eval-harness}
Leo Gao, Jonathan Tow, Baber Abbasi, Stella Biderman, Sid Black, Anthony DiPofi, Charles Foster, Laurence Golding, Jeffrey Hsu, Alain Le~Noac'h, Haonan Li, Kyle McDonell, Niklas Muennighoff, Chris Ociepa, Jason Phang, Laria Reynolds, Hailey Schoelkopf, Aviya Skowron, Lintang Sutawika, Eric Tang, Anish Thite, Ben Wang, Kevin Wang, and Andy Zou. 2023.
\newblock \href {https://doi.org/10.5281/zenodo.10256836} {A framework for few-shot language model evaluation}.

\bibitem[{Gunasekar et~al.(2023)Gunasekar, Zhang, Aneja, Mendes, Del~Giorno, Gopi, Javaheripi, Kauffmann, de~Rosa, Saarikivi et~al.}]{gunasekar2023textbooks}
Suriya Gunasekar, Yi~Zhang, Jyoti Aneja, Caio C{\'e}sar~Teodoro Mendes, Allie Del~Giorno, Sivakanth Gopi, Mojan Javaheripi, Piero Kauffmann, Gustavo de~Rosa, Olli Saarikivi, et~al. 2023.
\newblock Textbooks are all you need.
\newblock \emph{arXiv preprint arXiv:2306.11644}.

\bibitem[{Heafield(2011)}]{heafield2011kenlm}
Kenneth Heafield. 2011.
\newblock Kenlm: Faster and smaller language model queries.
\newblock In \emph{Proceedings of the sixth workshop on statistical machine translation}, pages 187--197.

\bibitem[{Hendrycks et~al.(2021)Hendrycks, Burns, Basart, Zou, Mazeika, Song, and Steinhardt}]{hendrycks2021measuring}
Dan Hendrycks, Collin Burns, Steven Basart, Andy Zou, Mantas Mazeika, Dawn Song, and Jacob Steinhardt. 2021.
\newblock \href {https://openreview.net/forum?id=d7KBjmI3GmQ} {Measuring massive multitask language understanding}.
\newblock In \emph{International Conference on Learning Representations}.

\bibitem[{Joulin et~al.(2016)Joulin, Grave, Bojanowski, Douze, J{\'e}gou, and Mikolov}]{joulin2016fasttext}
Armand Joulin, Edouard Grave, Piotr Bojanowski, Matthijs Douze, H{\'e}rve J{\'e}gou, and Tomas Mikolov. 2016.
\newblock Fasttext.zip: Compressing text classification models.
\newblock \emph{arXiv preprint arXiv:1612.03651}.

\bibitem[{Joulin et~al.(2017)Joulin, Grave, Bojanowski, and Mikolov}]{joulin2016bag}
Armand Joulin, Edouard Grave, Piotr Bojanowski, and Tomas Mikolov. 2017.
\newblock \href {https://aclanthology.org/E17-2068/} {Bag of tricks for efficient text classification}.
\newblock In \emph{Proceedings of the 15th Conference of the {E}uropean Chapter of the Association for Computational Linguistics: Volume 2, Short Papers}, pages 427--431, Valencia, Spain. Association for Computational Linguistics.

\bibitem[{Kudugunta et~al.(2023)Kudugunta, Caswell, Zhang, Garcia, Xin, Kusupati, Stella, Bapna, and Firat}]{kudugunta2023madlad400multilingualdocumentlevellarge}
Sneha Kudugunta, Isaac Caswell, Biao Zhang, Xavier Garcia, Derrick Xin, Aditya Kusupati, Romi Stella, Ankur Bapna, and Orhan Firat. 2023.
\newblock Madlad-400: A multilingual and document-level large audited dataset.
\newblock \emph{Advances in Neural Information Processing Systems}, 36:67284--67296.

\bibitem[{Lai et~al.(2017)Lai, Xie, Liu, Yang, and Hovy}]{lai2017race}
Guokun Lai, Qizhe Xie, Hanxiao Liu, Yiming Yang, and Eduard Hovy. 2017.
\newblock Race: Large-scale reading comprehension dataset from examinations.
\newblock In \emph{Proceedings of the 2017 Conference on Empirical Methods in Natural Language Processing}, pages 785--794.

\bibitem[{Lee et~al.(2022)Lee, Ippolito, Nystrom, Zhang, Eck, Callison-Burch, and Carlini}]{lee2022deduplicating}
Katherine Lee, Daphne Ippolito, Andrew Nystrom, Chiyuan Zhang, Douglas Eck, Chris Callison-Burch, and Nicholas Carlini. 2022.
\newblock Deduplicating training data makes language models better.
\newblock In \emph{Proceedings of the 60th Annual Meeting of the Association for Computational Linguistics (Volume 1: Long Papers)}, pages 8424--8445.

\bibitem[{Li et~al.(2024)Li, Fang, Smyrnis, Ivgi, Jordan, Gadre, Bansal, Guha, Keh, Arora et~al.}]{li2024datacomp}
Jeffrey Li, Alex Fang, Georgios Smyrnis, Maor Ivgi, Matt Jordan, Samir Gadre, Hritik Bansal, Etash Guha, Sedrick Keh, Kushal Arora, et~al. 2024.
\newblock \href {https://openreview.net/forum?id=CNWdWn47IE} {Datacomp-{LM}: In search of the next generation of training sets for language models}.
\newblock In \emph{The Thirty-eight Conference on Neural Information Processing Systems Datasets and Benchmarks Track}.

\bibitem[{Li et~al.(2023)Li, Bubeck, Eldan, Del~Giorno, Gunasekar, and Lee}]{li2023textbooks}
Yuanzhi Li, S{\'e}bastien Bubeck, Ronen Eldan, Allie Del~Giorno, Suriya Gunasekar, and Yin~Tat Lee. 2023.
\newblock Textbooks are all you need ii: phi-1.5 technical report.
\newblock \emph{arXiv preprint arXiv:2309.05463}.

\bibitem[{Maini et~al.(2024)Maini, Seto, Bai, Grangier, Zhang, and Jaitly}]{maini2024rephrasing}
Pratyush Maini, Skyler Seto, He~Bai, David Grangier, Yizhe Zhang, and Navdeep Jaitly. 2024.
\newblock Rephrasing the web: A recipe for compute and data-efficient language modeling.
\newblock In \emph{ICLR 2024 Workshop on Navigating and Addressing Data Problems for Foundation Models}.

\bibitem[{Merrick et~al.(2024)Merrick, Xu, Nuti, and Campos}]{merrick2024arctic}
Luke Merrick, Danmei Xu, Gaurav Nuti, and Daniel Campos. 2024.
\newblock Arctic-embed: Scalable, efficient, and accurate text embedding models.
\newblock \emph{arXiv preprint arXiv:2405.05374}.

\bibitem[{Mihaylov et~al.(2018)Mihaylov, Clark, Khot, and Sabharwal}]{mihaylov2018can}
Todor Mihaylov, Peter Clark, Tushar Khot, and Ashish Sabharwal. 2018.
\newblock Can a suit of armor conduct electricity? a new dataset for open book question answering.
\newblock In \emph{Proceedings of the 2018 Conference on Empirical Methods in Natural Language Processing}, pages 2381--2391.

\bibitem[{Muennighoff et~al.(2024)Muennighoff, Rush, Barak, Le~Scao, Tazi, Piktus, Pyysalo, Wolf, and Raffel}]{muennighoff2024scaling}
Niklas Muennighoff, Alexander Rush, Boaz Barak, Teven Le~Scao, Nouamane Tazi, Aleksandra Piktus, Sampo Pyysalo, Thomas Wolf, and Colin~A Raffel. 2024.
\newblock Scaling data-constrained language models.
\newblock \emph{Advances in Neural Information Processing Systems}, 36.

\bibitem[{Parmar et~al.(2024)Parmar, Prabhumoye, Jennings, Liu, Jhunjhunwala, Wang, Patwary, Shoeybi, and Catanzaro}]{parmar2024data}
Jupinder Parmar, Shrimai Prabhumoye, Joseph Jennings, Bo~Liu, Aastha Jhunjhunwala, Zhilin Wang, Mostofa Patwary, Mohammad Shoeybi, and Bryan Catanzaro. 2024.
\newblock Data, data everywhere: A guide for pretraining dataset construction.
\newblock In \emph{Proceedings of the 2024 Conference on Empirical Methods in Natural Language Processing}, pages 10671--10695.

\bibitem[{Penedo et~al.(2024)Penedo, Kydl{\'\i}{\v{c}}ek, allal, Lozhkov, Mitchell, Raffel, Werra, and Wolf}]{penedo2024fineweb}
Guilherme Penedo, Hynek Kydl{\'\i}{\v{c}}ek, Loubna~Ben allal, Anton Lozhkov, Margaret Mitchell, Colin Raffel, Leandro~Von Werra, and Thomas Wolf. 2024.
\newblock \href {https://openreview.net/forum?id=n6SCkn2QaG} {The fineweb datasets: Decanting the web for the finest text data at scale}.
\newblock In \emph{The Thirty-eight Conference on Neural Information Processing Systems Datasets and Benchmarks Track}.

\bibitem[{Penedo et~al.(2023)Penedo, Malartic, Hesslow, Cojocaru, Alobeidli, Cappelli, Pannier, Almazrouei, and Launay}]{penedo2023refinedweb}
Guilherme Penedo, Quentin Malartic, Daniel Hesslow, Ruxandra Cojocaru, Hamza Alobeidli, Alessandro Cappelli, Baptiste Pannier, Ebtesam Almazrouei, and Julien Launay. 2023.
\newblock The refinedweb dataset for falcon llm: Outperforming curated corpora with web data only.
\newblock \emph{Advances in Neural Information Processing Systems}, 36:79155--79172.

\bibitem[{Pomik{\'a}lek(2011)}]{pomikalek2011removing}
Jan Pomik{\'a}lek. 2011.
\newblock Removing boilerplate and duplicate content from web corpora.
\newblock \emph{Disertacn{\i} pr{\'a}ce, Masarykova univerzita, Fakulta informatiky}.

\bibitem[{Rae et~al.(2021)Rae, Borgeaud, Cai, Millican, Hoffmann, Song, Aslanides, Henderson, Ring, Young et~al.}]{rae2021scaling}
Jack~W Rae, Sebastian Borgeaud, Trevor Cai, Katie Millican, Jordan Hoffmann, Francis Song, John Aslanides, Sarah Henderson, Roman Ring, Susannah Young, et~al. 2021.
\newblock Scaling language models: Methods, analysis \& insights from training gopher.
\newblock \emph{arXiv preprint arXiv:2112.11446}.

\bibitem[{Raffel et~al.(2020)Raffel, Shazeer, Roberts, Lee, Narang, Matena, Zhou, Li, and Liu}]{raffel2020exploring}
Colin Raffel, Noam Shazeer, Adam Roberts, Katherine Lee, Sharan Narang, Michael Matena, Yanqi Zhou, Wei Li, and Peter~J Liu. 2020.
\newblock Exploring the limits of transfer learning with a unified text-to-text transformer.
\newblock \emph{Journal of machine learning research}, 21(140):1--67.

\bibitem[{Sakaguchi et~al.(2021)Sakaguchi, Bras, Bhagavatula, and Choi}]{sakaguchi2021winogrande}
Keisuke Sakaguchi, Ronan~Le Bras, Chandra Bhagavatula, and Yejin Choi. 2021.
\newblock Winogrande: An adversarial winograd schema challenge at scale.
\newblock \emph{Communications of the ACM}, 64(9):99--106.

\bibitem[{Sap et~al.(2019)Sap, Rashkin, Chen, Le~Bras, and Choi}]{sap2019social}
Maarten Sap, Hannah Rashkin, Derek Chen, Ronan Le~Bras, and Yejin Choi. 2019.
\newblock Social iqa: Commonsense reasoning about social interactions.
\newblock In \emph{Proceedings of the 2019 Conference on Empirical Methods in Natural Language Processing and the 9th International Joint Conference on Natural Language Processing (EMNLP-IJCNLP)}, pages 4463--4473.

\bibitem[{Shazeer(2020)}]{shazeer2020glu}
Noam Shazeer. 2020.
\newblock Glu variants improve transformer.
\newblock \emph{arXiv preprint arXiv:2002.05202}.

\bibitem[{Shoeybi et~al.(2019)Shoeybi, Patwary, Puri, LeGresley, Casper, and Catanzaro}]{shoeybi2019megatron}
Mohammad Shoeybi, Mostofa Patwary, Raul Puri, Patrick LeGresley, Jared Casper, and Bryan Catanzaro. 2019.
\newblock Megatron-lm: Training multi-billion parameter language models using model parallelism.
\newblock \emph{arXiv preprint arXiv:1909.08053}.

\bibitem[{Soldaini et~al.(2024)Soldaini, Kinney, Bhagia, Schwenk, Atkinson, Authur, Bogin, Chandu, Dumas, Elazar et~al.}]{soldaini2024dolma}
Luca Soldaini, Rodney Kinney, Akshita Bhagia, Dustin Schwenk, David Atkinson, Russell Authur, Ben Bogin, Khyathi Chandu, Jennifer Dumas, Yanai Elazar, et~al. 2024.
\newblock Dolma: an open corpus of three trillion tokens for language model pretraining research.
\newblock In \emph{Proceedings of the 62nd Annual Meeting of the Association for Computational Linguistics (Volume 1: Long Papers)}, pages 15725--15788.

\bibitem[{Talmor et~al.(2019)Talmor, Herzig, Lourie, and Berant}]{talmor2019commonsenseqa}
Alon Talmor, Jonathan Herzig, Nicholas Lourie, and Jonathan Berant. 2019.
\newblock Commonsenseqa: A question answering challenge targeting commonsense knowledge.
\newblock In \emph{Proceedings of the 2019 Conference of the North American Chapter of the Association for Computational Linguistics: Human Language Technologies, Volume 1 (Long and Short Papers)}, pages 4149--4158.

\bibitem[{Tang et~al.(2024)Tang, Ranjan, Pangarkar, Liang, Wang, An, Rao, Jin, Wang, Cheng, Sun, Mu, Miller, Ma, Peng, Liu, and Xing}]{txt360data2024}
Liping Tang, Nikhil Ranjan, Omkar Pangarkar, Xuezhi Liang, Zhen Wang, Li~An, Bhaskar Rao, Linghao Jin, Huijuan Wang, Zhoujun Cheng, Suqi Sun, Cun Mu, Victor Miller, Xuezhe Ma, Yue Peng, Zhengzhong Liu, and Eric~P. Xing. 2024.
\newblock Txt360: A top-quality llm pre-training dataset requires the perfect blend.
\newblock \url{https://huggingface.co/spaces/LLM360/TxT360}.
\newblock Accessed: October 24, 2024.

\bibitem[{Team et~al.(2024)Team, Riviere, Pathak, Sessa, Hardin, Bhupatiraju, Hussenot, Mesnard, Shahriari, Ram{\'e} et~al.}]{team2024gemma}
Gemma Team, Morgane Riviere, Shreya Pathak, Pier~Giuseppe Sessa, Cassidy Hardin, Surya Bhupatiraju, L{\'e}onard Hussenot, Thomas Mesnard, Bobak Shahriari, Alexandre Ram{\'e}, et~al. 2024.
\newblock Gemma 2: Improving open language models at a practical size.
\newblock \emph{arXiv preprint arXiv:2408.00118}.

\bibitem[{Teknium(2023)}]{OpenHermes}
Teknium. 2023.
\newblock Openhermes 2.5: An open dataset of synthetic data for generalist llm assistants.
\newblock \url{https://huggingface.co/datasets/teknium/OpenHermes-2.5}.
\newblock Accessed: October 24, 2024.

\bibitem[{Tokpanov et~al.(2024)Tokpanov, Glorioso, Anthony, and Millidge}]{tokpanov2024zyda25trilliontoken}
Yury Tokpanov, Paolo Glorioso, Quentin Anthony, and Beren Millidge. 2024.
\newblock \href {https://arxiv.org/abs/2411.06068} {Zyda-2: a 5 trillion token high-quality dataset}.
\newblock \emph{Preprint}, arXiv:2411.06068.

\bibitem[{Wang et~al.(2023)Wang, Kordi, Mishra, Liu, Smith, Khashabi, and Hajishirzi}]{wang2023self}
Yizhong Wang, Yeganeh Kordi, Swaroop Mishra, Alisa Liu, Noah~A Smith, Daniel Khashabi, and Hannaneh Hajishirzi. 2023.
\newblock Self-instruct: Aligning language models with self-generated instructions.
\newblock In \emph{Proceedings of the 61st Annual Meeting of the Association for Computational Linguistics (Volume 1: Long Papers)}, pages 13484--13508.

\bibitem[{Wenzek et~al.(2020)Wenzek, Lachaux, Conneau, Chaudhary, Guzm{\'a}n, Joulin, and Grave}]{wenzek-etal-2020-ccnet}
Guillaume Wenzek, Marie-Anne Lachaux, Alexis Conneau, Vishrav Chaudhary, Francisco Guzm{\'a}n, Armand Joulin, and Edouard Grave. 2020.
\newblock \href {https://aclanthology.org/2020.lrec-1.494} {{CCN}et: Extracting high quality monolingual datasets from web crawl data}.
\newblock In \emph{Proceedings of the Twelfth Language Resources and Evaluation Conference}, pages 4003--4012, Marseille, France. European Language Resources Association.

\bibitem[{Xue et~al.(2021)Xue, Constant, Roberts, Kale, Al-Rfou, Siddhant, Barua, and Raffel}]{xue-etal-2021-mt5}
Linting Xue, Noah Constant, Adam Roberts, Mihir Kale, Rami Al-Rfou, Aditya Siddhant, Aditya Barua, and Colin Raffel. 2021.
\newblock \href {https://doi.org/10.18653/v1/2021.naacl-main.41} {m{T}5: A massively multilingual pre-trained text-to-text transformer}.
\newblock In \emph{Proceedings of the 2021 Conference of the North American Chapter of the Association for Computational Linguistics: Human Language Technologies}, pages 483--498, Online. Association for Computational Linguistics.

\bibitem[{Yang et~al.(2024)Yang, Yang, Hui, Zheng, Yu, Zhou, Li, Li, Liu, Huang et~al.}]{yang2024qwen2}
An~Yang, Baosong Yang, Binyuan Hui, Bo~Zheng, Bowen Yu, Chang Zhou, Chengpeng Li, Chengyuan Li, Dayiheng Liu, Fei Huang, et~al. 2024.
\newblock Qwen2 technical report.
\newblock \emph{arXiv preprint arXiv:2407.10671}.

\bibitem[{Zellers et~al.(2019)Zellers, Holtzman, Bisk, Farhadi, and Choi}]{zellers2019hellaswag}
Rowan Zellers, Ari Holtzman, Yonatan Bisk, Ali Farhadi, and Yejin Choi. 2019.
\newblock Hellaswag: Can a machine really finish your sentence?
\newblock In \emph{Proceedings of the 57th Annual Meeting of the Association for Computational Linguistics}, pages 4791--4800.

\end{thebibliography}

\newpage

\appendix

\section{Pipeline Overview}
\label{appendix:pipelineoverview}

\begin{figure*}[htbp]
\centering
\begin{tikzpicture}[
    node distance=0.5cm and 0.3cm, 
    doc/.style={shape=rectangle, draw, fill=orange!30, text centered, font=\scriptsize, inner sep=5pt},
    process/.style={shape=ellipse, draw, fill=green!10, text centered, text width=2cm, font=\scriptsize, inner sep=3pt},
    arrow/.style={-Stealth, thick},
    dataflow/.style={Stealth-Stealth, thick, dashed}
]

\node[doc] (ccdata) {Common Crawl data};
\node[process, below=of ccdata] (text_extraction) {Text extraction};
\node[process, below=of text_extraction] (lang_id) {Language identification};
\node[process, below=of lang_id] (deduplication) {Deduplication};
\node[process, below=of deduplication] (classifier_filtering) {Ensemble-of-classifier-based filtering};
\node[process, below=of classifier_filtering] (bucketing) {Bucketing};

\draw[arrow] (ccdata) -- (text_extraction);
\draw[arrow] (text_extraction) -- (lang_id);
\draw[arrow] (lang_id) -- node[right, pos=0.4, font=\tiny] {English} (deduplication);
\draw[arrow] (deduplication) -- (classifier_filtering);
\draw[arrow] (classifier_filtering) -- (bucketing);

\node[doc, below=of bucketing] (medium_quality_label) {Medium quality data};
\node[process, below=of medium_quality_label] (heuristic_filtering_2) {Heuristic filtering};
\node[doc, below=1.5cm of heuristic_filtering_2] (final_dataset) {Final dataset};

\node[doc, left=2cm of medium_quality_label] (low_quality_label) {Low quality data};
\node[process, below=of low_quality_label] (heuristic_filtering) {Heuristic filtering};
\node[process, below=of heuristic_filtering] (sdg_wiki) {SDG (Wikipedia-style rephrasing)};

\node[doc, right=2cm of medium_quality_label] (high_quality_label) {High quality data};
\node[process, below=of high_quality_label] (sdg_diverse) {SDG (Diverse QA, Distillation, Knowledge extraction, Knowledge listing, Wikipedia-style rephrasing)};

\draw[arrow] (bucketing) -- (medium_quality_label);
\draw[arrow] (medium_quality_label) -- (heuristic_filtering_2);
\draw[arrow] (heuristic_filtering_2) -- (final_dataset);

\draw[arrow] (bucketing) -- (low_quality_label);
\draw[arrow] (low_quality_label) -- (heuristic_filtering);
\draw[arrow] (heuristic_filtering) -- (final_dataset);
\draw[arrow] (heuristic_filtering) -- (sdg_wiki);
\draw[arrow] (sdg_wiki) -- (final_dataset);

\draw[arrow] (bucketing) -- (high_quality_label);
\draw[arrow] (high_quality_label) -- (final_dataset);
\draw[arrow] (high_quality_label) -- (sdg_diverse);
\draw[arrow] (sdg_diverse) -- (final_dataset);

\end{tikzpicture}
\caption{Pipeline overview}
\label{fig:pipelineoverview}
\end{figure*}

An overview of the pipeline is shown in Figure~\ref{fig:pipelineoverview}.

\section{Comparison of FineWeb-Edu and DCLM Classifier}
\label{sec:appendix-1}
Different classifiers have different standards for high-quality documents. Thus, ensemble multiple classifiers will help increase the recall of high-quality documents. We did a detailed comparison of two of the classifiers that we employ in our method: the FineWeb-Edu classifier which score document quality based on their educational-level, and the DCLM based classifier which value the informativeness of the document. 

We compare the high-quality documents predicted by the two classifiers on one Common Crawl snapshot (dated 2021-21). Table~\ref{table:hq-doc-comparision} show the document statistics comparison. We further show the detailed URL domains comparison between the two classifiers' predictions in Table~\ref{table:ap-domain-comparision}. We can see that each classifier has their own high-quality domain preferences. Among the top 1k domains, only 368 domains are in the intersection. Therefore, ensemble of different classifiers can help increase retrieving more high-quality documents from Common Crawl.

\begin{table*}[thb]
\centering
\resizebox{\textwidth}{!}{
\begin{tabular}{|lclclll|}
\hline
\multicolumn{7}{|c|}{\textbf{Top domains and domain overlap analysis =\textgreater 368 domains are in top 1k domains of both}} \\ \hline
\multicolumn{1}{|c}{} & \multicolumn{1}{c|}{} & \multicolumn{1}{c}{} & \multicolumn{1}{c|}{} & \multicolumn{3}{c|}{\textbf{Top 1k Domains}} \\ \cline{5-7} 
\multicolumn{1}{|c}{\multirow{-2}{*}{\textbf{\begin{tabular}[c]{@{}c@{}}FineWeb-Edu \\ Top Domains\end{tabular}}}} & \multicolumn{1}{c|}{\multirow{-2}{*}{\textbf{Count}}} & \multicolumn{1}{c}{\multirow{-2}{*}{\textbf{\begin{tabular}[c]{@{}c@{}}DCLM\\ Top Domains\end{tabular}}}} & \multicolumn{1}{c|}{\multirow{-2}{*}{\textbf{Count}}} & \multicolumn{1}{l|}{Intersection (368)} & \multicolumn{1}{l|}{In FineWeb-Edu only} & In DCLM only \\ \hline
wordpress.com & \multicolumn{1}{c|}{39228} & wordpress.com & \multicolumn{1}{c|}{85378} & \multicolumn{1}{l|}{{ {123helpme.com}}} & \multicolumn{1}{l|}{{ {111papers.com}}} & { {4archive.org}} \\
thefreedictionary.com & \multicolumn{1}{c|}{20420} & stackexchange.com & \multicolumn{1}{c|}{64831} & \multicolumn{1}{l|}{{ {24houranswers.com}}} & \multicolumn{1}{l|}{{ {3dprint.com}}} & { {4channel.org}} \\
stackexchange.com & \multicolumn{1}{c|}{17853} & livejournal.com & \multicolumn{1}{c|}{36521} & \multicolumn{1}{l|}{{ {abc.net.au}}} & \multicolumn{1}{l|}{{ {aafp.org}}} & { {4hw.com.cn}} \\
britannica.com & \multicolumn{1}{c|}{14761} & medium.com & \multicolumn{1}{c|}{27347} & \multicolumn{1}{l|}{{ {abovetopsecret.com}}} & \multicolumn{1}{l|}{{ {aappublications.org}}} & { {5winebar.com}} \\
ipl.org & \multicolumn{1}{c|}{13132} & fandom.com & \multicolumn{1}{c|}{13986} & \multicolumn{1}{l|}{{ {academickids.com}}} & \multicolumn{1}{l|}{{ {abs.gov.au}}} & { {aawsat.com}} \\
medium.com & \multicolumn{1}{c|}{11539} & ipl.org & \multicolumn{1}{c|}{12282} & \multicolumn{1}{l|}{{ {adafruit.com}}} & \multicolumn{1}{l|}{{ {accessgenealogy.com}}} & { {abc11.com}} \\
nih.gov & \multicolumn{1}{c|}{10624} & answers.com & \multicolumn{1}{c|}{10790} & \multicolumn{1}{l|}{{ {adobe.com}}} & \multicolumn{1}{l|}{{ {achrnews.com}}} & { {abc30.com}} \\
igi-global.com & \multicolumn{1}{c|}{9136} & nih.gov & \multicolumn{1}{c|}{9091} & \multicolumn{1}{l|}{{ {alchetron.com}}} & \multicolumn{1}{l|}{{ {acm.org}}} & { {abc7chicago.com}} \\
slideplayer.com & \multicolumn{1}{c|}{8460} & typepad.com & \multicolumn{1}{c|}{8078} & \multicolumn{1}{l|}{{ {aljazeera.com}}} & \multicolumn{1}{l|}{{ {adidasshoesoutletwholesale.com}}} & { {able2know.org}} \\
answers.com & \multicolumn{1}{c|}{8103} & commonsensemedia.org & \multicolumn{1}{c|}{7772} & \multicolumn{1}{l|}{{ {allegancountyedc.com}}} & \multicolumn{1}{l|}{{ {adslspeedtest.net}}} & { {aceshowbiz.com}} \\
wikipedia.org & \multicolumn{1}{c|}{6867} & wsj.com & \multicolumn{1}{c|}{7652} & \multicolumn{1}{l|}{{ {allinterview.com}}} & \multicolumn{1}{l|}{{ {aero-net.org}}} & { {activerain.com}} \\
dictionary.com & \multicolumn{1}{c|}{6763} & imdb.com & \multicolumn{1}{c|}{7263} & \multicolumn{1}{l|}{{ {amazon.com}}} & \multicolumn{1}{l|}{{ {agwired.com}}} & { {addicted2success.com}} \\
en-academic.com & \multicolumn{1}{c|}{5292} & theatlantic.com & \multicolumn{1}{c|}{7008} & \multicolumn{1}{l|}{{ {americanbar.org}}} & \multicolumn{1}{l|}{{ {ahdictionary.com}}} & { {additudemag.com}} \\
sciencemag.org & \multicolumn{1}{c|}{5254} & yahoo.com & \multicolumn{1}{c|}{5921} & \multicolumn{1}{l|}{{ {angelfire.com}}} & \multicolumn{1}{l|}{{ {ajol.info}}} & { {agingcare.com}} \\
brainscape.com & \multicolumn{1}{c|}{5129} & fanfiction.net & \multicolumn{1}{c|}{5499} & \multicolumn{1}{l|}{{ {answers.com}}} & \multicolumn{1}{l|}{{ {akjournals.com}}} & { {agnostic.com}} \\
encyclopedia.com & \multicolumn{1}{c|}{4698} & huffpost.com & \multicolumn{1}{c|}{5471} & \multicolumn{1}{l|}{{ {antiessays.com}}} & \multicolumn{1}{l|}{{ {aleteia.org}}} & { {airmilescalculator.com}} \\
nasa.gov & \multicolumn{1}{c|}{4615} & adobe.com & \multicolumn{1}{c|}{5182} & \multicolumn{1}{l|}{{ {apple.com}}} & \multicolumn{1}{l|}{{ {alison.com}}} & { {airportia.com}} \\
slideserve.com & \multicolumn{1}{c|}{4538} & scribd.com & \multicolumn{1}{c|}{4948} & \multicolumn{1}{l|}{{ {archive.org}}} & \multicolumn{1}{l|}{{ {all-creatures.org}}} & { {alarabiya.net}} \\
scribd.com & \multicolumn{1}{c|}{4430} & thefreedictionary.com & \multicolumn{1}{c|}{4847} & \multicolumn{1}{l|}{{ {arduino.cc}}} & \multicolumn{1}{l|}{{ {allaboutheaven.org}}} & { {alex-in-wonderland.com}} \\
kiddle.co & \multicolumn{1}{c|}{4323} & mathworks.com & \multicolumn{1}{c|}{4655} & \multicolumn{1}{l|}{{ {arstechnica.com}}} & \multicolumn{1}{l|}{{ {allthatsinteresting.com}}} & { {alexa-gueguen.com}} \\ \hline
\end{tabular}
}
\caption{High Quality Documents Domains Comparison. 368 Top Domains are in the intersection.}
\label{table:ap-domain-comparision}
\end{table*}

\section{Bucket Comparison} \label{appendix:bucketcomparison}

\begin{figure}[htbp]
\centering
\includegraphics[width=0.4\textwidth]{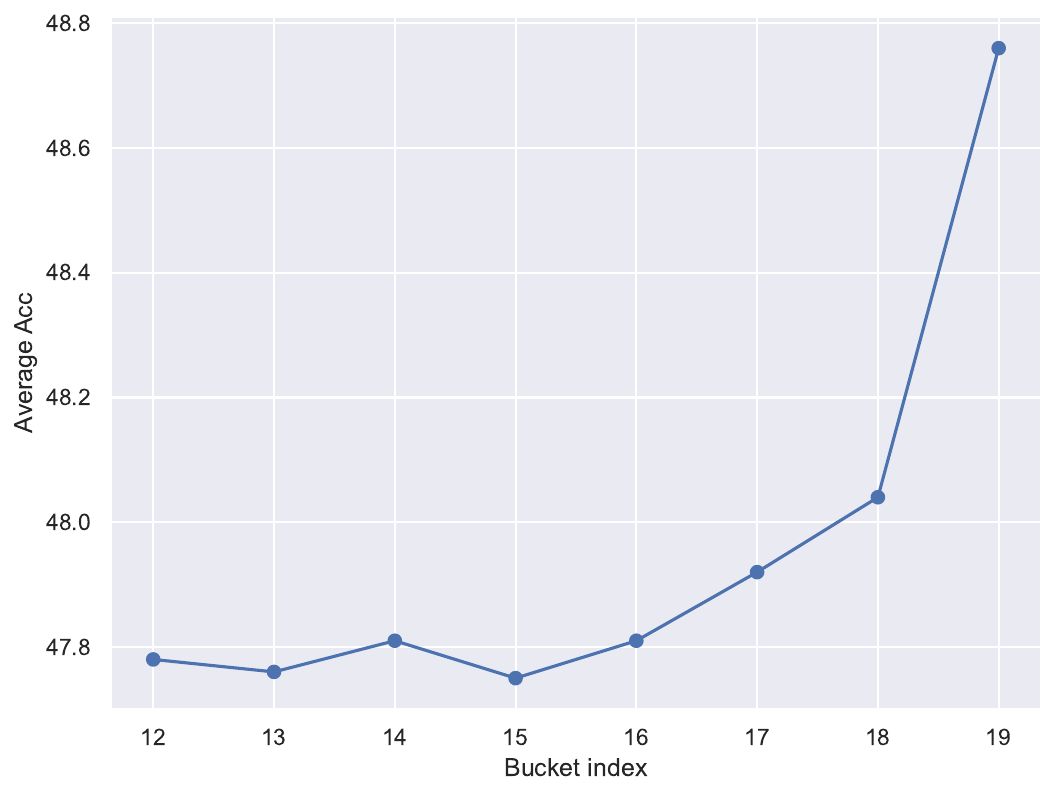}
\caption{Ablation study on the buckets.}
\label{fig:ablation_buckets}
\end{figure}

To better understand the quality of data in each of our 20 data buckets, we carry out ablation studies to test their benchmark accuracies. For each study, we take a 900B-token checkpoint and continue the pretraining for 50B more tokens. For 34\% of the 50B tokens we used the bucket data being tested, while we fixed the other 66\% as the same data distribution of the 900B pretraining process to make sure the distribution did not shift too much. See Figure~\ref{fig:ablation_buckets} for the results. The average accuracy is calculated across 13 downstream tasks. Note that Bucket 19 greatly outperforms all other buckets and the differences within bucket 12-18 are marginal. We used the results here as a reference when designing the quality labels in Table~\ref{table:quality_label_stats}. 

\section{Training Details: Ablations}
\label{appendix:hyperparameters}
\begin{table}[!htbp] \small \centering
\begin{tabularx}{\linewidth}{Xr}
\toprule
\textbf{Category}    & \textbf{Blend \%} \\ \midrule
English Common Crawl & 73                \\
Books and patents    & 9                 \\
Papers               & 9                 \\
Code                 & 5                 \\
Conversational       & 3                 \\
Wikipedia            & 1                 \\ \bottomrule
\end{tabularx}
\caption{Data blend for the experiments with 8B parameter transformer LLMs trained for 1T tokens. Experiments in this paper varied only the $73\%$ English Common Crawl portion.}
\label{table:datablend}
\end{table}

As mentioned in Section~\ref{section:experiment-setup},
we use the open source Megatron-LM library\footnote{\url{https://github.com/NVIDIA/Megatron-LM}}~\citep{shoeybi2019megatron} to train 8B parameter transformer LLMs for 1T tokens. The key hyperparameters are as follows: We use 32 transformer layers with hidden dimension 4096, 32 attention heads, and SwiGLU activations~\cite{shazeer2020glu}. For the attention, we use grouped query attention with 8 query groups~\citep{ainslie2023gqa}. We use the Adam optimizer with $\beta_1 = 0.9, \beta_2 = 0.95, \epsilon=1e{-8}$, weight decay $0.1$, and the cosine learning rate schedule with peak learning rate at 3e-4 and minimum learning rate at 3e-6. A single training run takes about 40 hours using 1024 NVIDIA H100 GPUs.

The data blend breakdown for these experiments is shown in Table~\ref{table:datablend}. Experiments in this paper varied only the $73\%$ English Common Crawl portion.

\section{Long-Horizon Curriculum Details}
\label{appendix:8b15t}

For the 15T token training run, a two-phase curriculum was employed that is described in more detail in~\citet{feng2024maximizedataspotentialenhancing}. The first phase of 9T tokens used $59\%$ English Common Crawl data (5.31T) and the second phase of 6T tokens used $31\%$ (1.86T), for a combined total of $47.8\%$ (7.17T). In the first phase, we used medium, medium-high, and high quality data (real and synthetic), and in the second phase we used only high quality data (real and synthetic).

\section{Common Crawl Snapshots} \label{sec:13_snapshots}

For the main datasets, we used the 99 snapshots CC-MAIN-2013-20 through CC-MAIN-2024-30.

The thirteen Common Crawl snapshots we use in some of the analysis and 1T token experiments are 
CC-MAIN-2023-23,     CC-MAIN-2023-14,     CC-MAIN-2023-06,     CC-MAIN-2022-49,     CC-MAIN-2022-27,     CC-MAIN-2022-05,     CC-MAIN-2021-43,     CC-MAIN-2021-21,     CC-MAIN-2021-04,     CC-MAIN-2020-45,     CC-MAIN-2020-29,     CC-MAIN-2020-05,     CC-MAIN-2019-35.

\section{Extractor \& Filter Ablation} \label{sec:non_mmlu}

The Avg tasks include ARC-Easy, ARC-Challenge, Hellaswag, Winogrande, RACE, PIQA, Commonsense QA, Openbook QA. 

Note that we only use FineWeb-Edu classifier for the quality labels of this ablation study and analysis. We do not use it in the final preparation of our dataset. See Section \ref{sec:classifier} for the details of our classifiers being used eventually to prepare the data.

\section{Prompt Templates}
\label{sec:appendix-prompt-templates}

Prompts 1-5 show the prompt templates we use for synthetic data generation.

\begin{table*}[!hbt]
\begin{prompt}{Prompt template: Diverse QA pairs}
Task: Read the text, ask questions and answer them.

Follow these instructions:
1. Ask diverse questions that require different cognitive skills or cover different aspects of the text.
2. Ask questions in various forms such as:
  - Yes/No questions that require determining whether a statement is true or false.
  - Open-ended questions that begin with words like what, how, when, where, why and who.
  - Multi-choice questions that offers two or more options to choose from. Include the options in the question.
  - Comparison questions that compare two quantities or objects and determine the relationship between them.
  - Reading comprehension questions that test the ability to understand and analyze the text.
  - Problem-solving questions that test the ability to solve mathematical, physical, or logical problems.
3. Focus on asking questions about factual information, important knowledge, or concrete details in the text.
4. Write questions and answers using clear and concise language.
5. Use plain text. Do not use Markdown.
6. Each question and answer pair should be on a separate line. Tag the question with "Question:" and the answer with "Answer:".

Text:
[DOCUMENT SEGMENT]

Task:
After reading the above text, ask up to 8 questions and provide the correct answers following the instructions. Give your response in this format:

Here are the questions and answers based on the provided text:
- Question: [first question] Answer: [first answer]
- Question: [second question] Answer: [second answer]
....
\end{prompt}
\end{table*}

\begin{table*}[!hbt]
\begin{prompt}{Prompt template: Distill.}
Your task is to read and paraphrase the provided text following these instructions:
- Aim to create a condensed but accurate and informative version of the original text, not a simplistic summary.
- Capture and preserve the crucial information, key concepts, important values, and factual details in the original text, while making it more readable and accessible.
- Retain technical terms, specialized vocabulary, and complex concepts.
- Retain examples, explanations of reasoning processes, and supporting evidence to maintain the text's depth and context.
- Only include information that is present in the original text. Do not adding new or unsubstantiated claims.
- Write in plain text.

Here is the text:
[DOCUMENT SEGMENT]

Task:
After thoroughly reading the above text, paraphrase it in high-quality and clear English following the instructions.
\end{prompt}
\end{table*}

\begin{table*}[!hbt]
\begin{prompt}{Prompt template: Knowledge list.}
Review the text and extract the key information. Follow these instructions:
- Carefully read the above text and provide a concise and organized list of factual information, concrete details, key concepts, and important numbers and statistics extracted from the text.
- Ensure each point is clear, specific, and supported by the original text.
- Ensure the extract text is information-dense and easier to learn from.
- Do not add titles or headings.

Text:
[DOCUMENT SEGMENT]

Task:
Extract the factual information, concrete details, and key concepts from the above text following the instructions.
\end{prompt}

\end{table*}

\begin{table*}[!hbt]
\begin{prompt}{Prompt template: Extract knowledge.}
Your task is to rewrite knowledge from the provided text following these instructions:
- Rewrite the text as a passage or passages using easy-to-understand and high-quality English like sentences in textbooks and Wikipedia.
- Focus on content in disciplines such as humanities, social sciences, natural sciences, technology, engineering, math, law and legal, business, management, art, education, agricultural sciences, politics, and history.
- Disregard content that does not contain useful facts or knowledge.
- Retain examples, explanations of reasoning processes, and supporting evidence to maintain the text's depth and context.
- Do not add or alter details. Only restate what is already in the text.
- Write in plain text.
- Do not add titles, subtitles, note, or comment.

Text:
[DOCUMENT SEGMENT]

Task:
Rewrite facts and knowledge from the above text as a passage or passages following the instructions.
\end{prompt}
\end{table*}

\begin{table*}[!hbt]
\begin{prompt}{Prompt template: Wikipedia-style rephrasing~\citep{maini2024rephrasing}.}
For the following paragraph give me a diverse paraphrase of the same in high quality English language as in sentences on Wikipedia. Begin your answer on a separate line with "Here is a paraphrased version:".

Text: [DOCUMENT SEGMENT]
\end{prompt}
\end{table*}

\end{document}